\documentclass[10pt, a4paper]{article}
\usepackage{tcolorbox}
\usepackage{subcaption}
\usepackage{wrapfig}
\usepackage{lrec-coling2024} 

\usepackage{hyperref}       
\usepackage{url}            
\usepackage{booktabs}       
\usepackage{amsfonts}       
\usepackage{nicefrac}       
\usepackage{microtype}      
\usepackage{xcolor}         
\usepackage{multirow}

\definecolor{beaublue}{rgb}{0.74, 0.83, 0.9}
\newenvironment{myblock}{%
  \begin{tcolorbox}[float,colback=beaublue!30!white,colframe=beaublue!50!black,title=Prompt for Tree-Instruct]
}{%
  \end{tcolorbox}
}

\title{Tree-Instruct: A Preliminary Study of the Intrinsic Relationship between Complexity and Alignment}

\name{Yingxiu Zhao$^1$, Bowen Yu$^2$\thanks{\small Correspondence to: Bowen Yu, Yongbin Li.}, Binyuan Hui$^2$, Haiyang Yu$^2$, Minghao Li$^2$, \\ \large \textbf{Fei Huang$^2$, Nevin L. Zhang$^1$, Yongbin Li$^2$}}

\address{The Hong Kong University of Science and Technology, Alibaba Group \\
         \texttt{\{yzhaocx,lzhang\}@connect.ust.hk}\\
        \texttt{\{yubowen.ybw,binyuan,yifei,liminghao,f.huang,shuide\}@alibaba-inc.com}
}

\abstract{Training large language models (LLMs) with open-domain instruction data has yielded remarkable success in aligning to end tasks and human preferences. Extensive research has highlighted the importance of the quality and diversity of instruction data. However, the impact of data complexity, as a crucial metric, remains relatively unexplored from three aspects: (1)where the sustainability of performance improvements with increasing complexity is uncertain; (2)whether the improvement brought by complexity merely comes from introducing more training tokens; and (3)where the potential benefits of incorporating instructions from easy to difficult are not yet fully understood. In this paper, we propose \textit{Tree-Instruct} to systematically enhance the instruction complexity in a controllable manner. By adding a specified number of nodes to instructions' semantic trees, this approach not only yields new instruction data from the modified tree but also allows us to control the difficulty level of modified instructions. Our preliminary experiments reveal the following insights: (1)Increasing complexity consistently leads to sustained performance improvements of LLMs. (2)Under the same token budget, a few complex instructions outperform diverse yet simple instructions. (3)Curriculum instruction tuning might not yield the anticipated results; focusing on increasing complexity appears to be the key.
 \\ \newline  }

\begin{document}

\maketitleabstract

\section{Introduction}
\label{sec:intro}
The latest generation of large language models (LLMs) has attracted significant attention due to their immense potential in language technologies~\cite{openai2022chatgpt,touvron2023llama,wei2023polylm,li2023api}. 
To enhance interactive user requests and chat interfaces, these models undergo instruction-tuning using supervised input-output pairs~\cite{iyer2022opt,jang2023exploring,chung2022scaling}. 
This process enables the model to comprehend the required style and format for effective user interaction, showcasing the knowledge and capabilities gained during pre-training~\cite{ouyang2022training}.

Consequently, the efficacy of instruction data significantly influences LLMs' abilities, shaping users' perceptions of their capabilities~\cite{wang2023aligning,kopf2023openassistant,chiang2023vicuna}. 
Recently, LIMA has demonstrated that with just 1000 carefully curated prompts and responses, an LLM can achieve remarkably strong performance~\cite{zhou2023lima}. This suggests that the scaling laws of instruction tuning are not solely dependent on data quantity but rather influenced by prompt diversity and quality.
However, one critical and less-explored aspect of evaluating instruction data is complexity. There are at least three unanswered questions related to complexity:
(1) \textit{\textbf{The scaling law of instruction complexity}: Intuitively, more complex instruction data might elicit more potential capabilities in LLMs to address intricate problems~\cite{luo2023wizardcoder,mukherjee2023orca}}. 
WizardLM~\cite{xu2023wizardlm} introduce in-depth and in-breadth evolving methods to rewrite prompts into more complex and diverse versions, resulting in a 12.4\% increase in LLMs' win rate with the same amount of data. 
Yet, whether WizardLM's performance improvement is due to complexity or merely derived from diversity remains uncertain. Moreover, the ongoing enhancements in complexity are yet to be explored.
(2) \textit{\textbf{The relationship between complexity-induced performance improvement and token quantity}}: Enhancing instance complexity inevitably increases the number of tokens per instance~\cite{dai2021preview}. While WizardLM exhibits performance improvements with the same instance quantity, it increases the number of tokens per instance. 
This raises the question of whether complexity-induced improvement in LLMs results from increased training tokens. As known, enlarging LLMs' pretraining token counts can lead to better performance~\cite{muennighoff2023scaling,tay2022transcending}.
(3) \textit{\textbf{The effectiveness of complexity-based curriculum instruction learning}}: Curriculum learning is a strategy in machine learning that starts with easy instances and gradually introduces harder ones~\cite{bengio2009curriculum}. 
Its effectiveness has been demonstrated in various NLP tasks like machine translation~\cite{zhou2020uncertainty}, dialogue systems~\cite{zhu2021combining}, and question answering~\cite{sachan2016easy}. However, its potential efficacy in instruction tuning is still under-explored.

However, to answer the aforementioned questions, the key hurdle lies in finding a controlled way to increase the complexity of instruction data without introducing unwanted factors such as diversity.
WizardLM~\cite{xu2023wizardlm} employs an in-depth evolving prompt like ``\textit{\texttt{Your objective is to rewrite a given prompt into a more complex version to make ChatGPT and GPT4 a bit harder to handle.}}'' to complicate the existing instructions.
Unfortunately, although intended to enhance complexity, this approach might inadvertently introduce diversity by diverting from the initial instruction objectives.
This issue becomes particularly severe when repeatedly employing in-depth evolving to achieve varying levels of complexity.
We study and analyze the instructions before and after in-depth evolving in Sec.~\ref{sec:consist}. 
As illustrated in Fig.~\ref{fig:inst}, the iteratively evolved instructions append additional objectives that deviate from the original instructions, showcasing a greater diversity. 

\begin{figure*}
    \centering
    \includegraphics[width=0.9\textwidth]{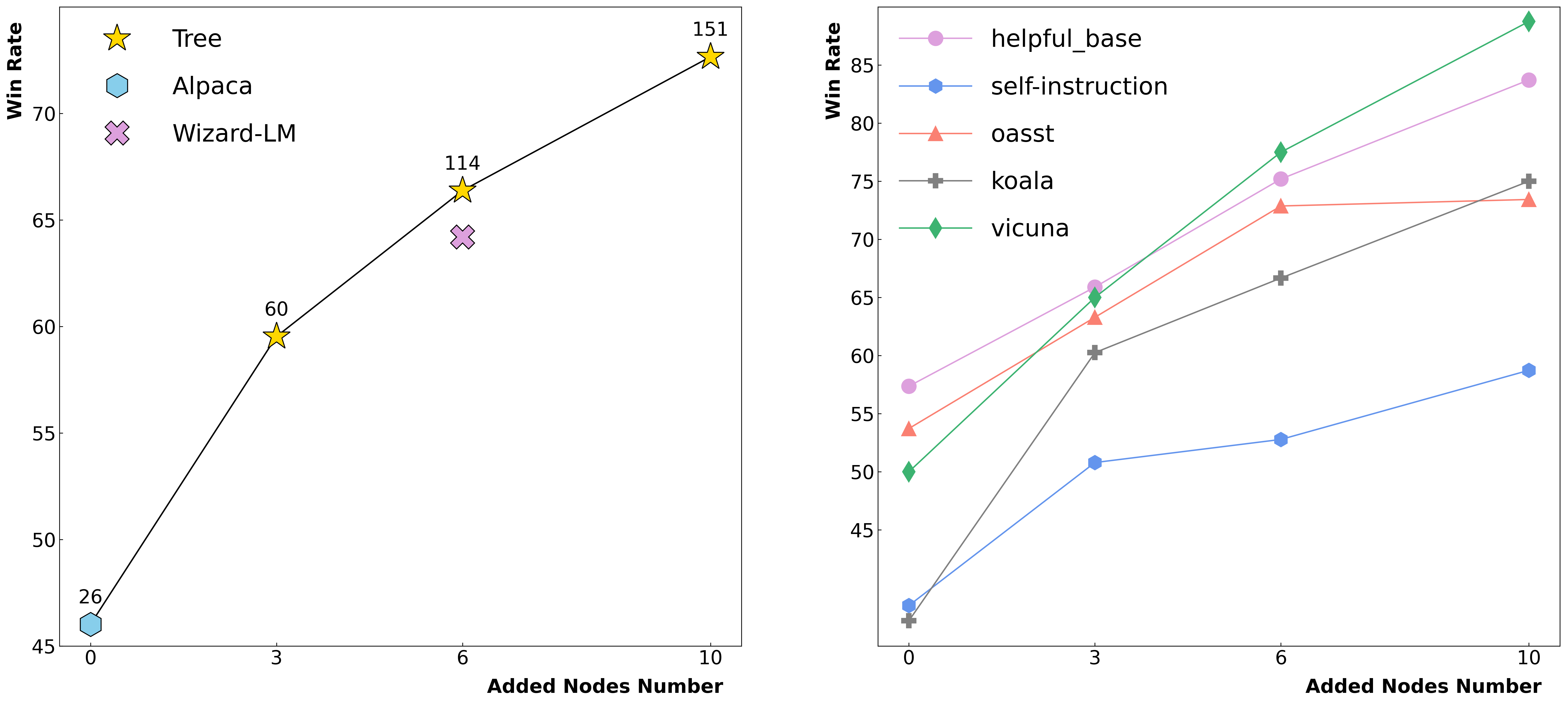}
    \caption{The scaling law of instruction complexity. We experiment with enhancing the complexity of semantic trees for 1,000 Alpaca instructions by adding extra 3, 6, and 10 nodes. We then evaluate models fine-tuned on instruction data of varying complexities against $\mathtt{text}$-$\mathtt{davinci003}$ in terms of win rate on AlpacaEval (Left). Additionally, we examine win rates on different subsets of AlpacaEval (Right). In the left figure, we indicate the average token count for instructions of different complexity levels. We also use WizardLM's in-depth deepening as the baseline.}
    \label{fig:scaling_law}
\end{figure*}

To address this concern, we propose \textbf{\textit{Tree-Instruct}}, which involves prompting LLMs to add a specific number of new nodes to the semantic tree of an existing instruction, as opposed to manipulating the text sequence directly, as done in Self-Instruct~\cite{wang2022self} or WizardLM~\cite{xu2023wizardlm}. 
We use the number of added nodes to represent the introduced level of complexity.
The advantage of this approach lies in the fact that semantic tree nodes lack any sequential order~\cite{shiv2019novel}. 
By enforcing LLMs to operate on the semantic tree, this process becomes analogous to inserting new words into the middle of the original instructions. 
This compels the models to complicate while adhering to the structural constraints of the initial instruction rather than merely appending new instructions. It can significantly mitigate the issue of straying from the primary theme of the initial instruction.
We leverage GPT-4 to assess the consistency of evolved instructions with original ones, and the results verify that Tree-Instruct improves WizardLM's consistency score from 0.56 to 0.69.
Fig.~\ref{fig:scaling_law} highlights how the number of added nodes raises the complexity level of the samples.

With the help of Tree-Instruct, we have obtained the following preliminary experimental conclusions:

(1) \textit{\textbf{As the complexity of the instruction data increases, the benefits of instruction tuning continue to grow}}: Following LIMA\citep{zhou2023lima}, we attempt instruction tuning using 1,000 samples from Alpaca-GPT-4 as a base. We add 3, 6, and 10 nodes to the semantic tree of each sample, resulting in performance gains of 13\%, 20\%, and 26\%, respectively, across eight sub-skills such as commonsense, writing, and coding, showing consistent improvements. 
Furthermore, this scaling law can be extended to more complex instruction data. For instance, when fine-tuning around 6,000 conversations filtered from ShareGPT via OpenChat\cite{openchat} (showing excellent performance in the open-source LLMs), we observe that by increasing the complexity through Tree-Instruct to around 1,100 users' instructions, the winning rate increases from 84.56\% to 86.19\% for AlpacaEval benchmark\footnote{\url{https://github.com/tatsu-lab/alpaca_eval}}.

(2) \textit{\textbf{The increase in complexity partly comes from additional tokens, but a few complex instructions outperform diverse yet simple instructions, under the same token budget.}}: We find that as the complexity increases, the number of tokens also increases. Adding ten nodes in the semantic tree increases the average token length of instructions from 186 to 607. 
Hence, to make a fair comparison, we increase the number of original instructions from 1,000 to 4,000 to match the total token quantity of our tree-instructed samples. 
Under this setting, the performance gain from adding ten nodes still achieves more than 15\%.
This indicates that the improvement due to complexity is partly attributed to the increased tokens, but increasing the complexity of samples is equivalent to the diversity achieved by four times the token count of simple samples. 
Moreover, with an equal number of instruction tokens, the evolution of Tree-Instruct yields a 2\% higher win rate than those from three iterations of in-depth evolution of WizardLM, demonstrating the superior efficacy of Tree-Instruct in enhancing complexity.

(3) \textit{\textbf{Curriculum instruction tuning may not be effective; increasing complexity is all you need}}: 
We implement curriculum learning by progressively training the LLM on increasing levels of difficulty. Initially, we train it on easy-level data with an addition of three nodes, followed by medium-level data with six nodes, and ultimately on hard data with ten nodes added.
 We observe that, when given the same number of training steps, curriculum learning does outperform training with a mixed difficulty of samples but still falls short compared to training solely on the added ten-node instruction data. This indicates that, as sample complexity increases, the significance of simpler samples diminishes significantly, suggesting that repeating training with complex samples may be sufficient.

\section{Related Work}
Large Language Models (LLMs), trained on extensive textual datasets, have risen as premier solutions for various NLP tasks~\cite{zhao2023survey}. 
Despite their remarkable performance, these models are not without their limitations. 
These limitations encompass potential misunderstandings of human instructions, the propensity to generate biased content, and the sporadic generation of hallucinated information. 
Consequently, bringing LLMs in line with human expectations has become a central focal point within the research community~\cite{bai2022training,song2023preference}.

To attain this alignment, researchers need to amass high-quality instructional data that authentically mirrors human needs and expectations. 
A rational starting point for data collection involves the adaptation of existing NLP benchmarks into natural language instructions, like PromptSource~\cite{bach2022promptsource}, SuperNaturalInstruction~\cite{wang2022super}, Unnatural Instructions~\cite{honovich2022unnatural} and FLAN~\cite{longpre2023flan} are spearheading this strategy. 
These benchmarks encompass a wide range of NLP tasks, spanning dialogue, reasoning, and coding, all unified under the realm of language instructions. 
TÜLU\cite{wang2023far} showcases that instructions from NLP tasks significantly bolster the reasoning prowess of aligned LLMs, where the diversity of tasks plays a pivotal role in shaping the capabilities of LLMs.

Nevertheless, a notable trend in NLP datasets is their propensity to emphasize particular skills, consequently yielding instructions that possess a somewhat confined scope. 
This constraint has the potential to impede their capacity to meet the intricate requirements of real-world applications. 
In order to tackle these challenges, one possible approach is to formulate instructions via purposeful human annotations.
An exemplary precursor to such a corpus is OpenAssistant~\cite{kopf2023openassistant}, which comprises over 10k dialogues involving the participation of 13k annotators from around the world. 
Another remarkable venture into harnessing human-generated instructions through crowd-sourcing is ShareGPT~\footnote{\url{2https://sharegpt.com/}}. This platform encourages users to contribute and exchange their engaging conversations with ChatGPT and GPT4.

While human annotation ensures both quality and diversity, it becomes challenging to ensure the quantity and complexity of instructional data due to the highly expensive annotation process~\cite{chen2023phoenix}, and the distribution of difficulty levels in human-created instructions tends to skew towards being either easy~\cite{luo2023wizardcoder}.
To address this issue, Self-Instruct~\cite{wang2022self} leverages ChatGPT's in-context learning capability to generate a large volume of instructions from a predefined set of human-annotated instructions spanning diverse topics and task types. 
Building upon this foundation, LIMA~\cite{zhou2023lima} and Alpagasus~\cite{chen2023alpagasus} separately validate the significant impact of data diversity and quality on instructional effectiveness. 
The selection of thousands of high-quality and diverse instructional examples proves more advantageous in achieving better results compared to using the entire dataset.
Further increasing the number of instructions could potentially induce a semantic shift in the LLMs~\cite{alshikh2023becoming}.
Up to this point, three key metrics within the instructional data—diversity, quality, and quantity—have been elucidated for their impact on tuning, though exploration into complexity remains insufficient. 
While WizardLM~\cite{xu2023wizardlm} demonstrates that evolving both the complexity and diversity of instructions can lead to performance enhancement, it does not deeply investigate the individual importance of complexity.
This paper introduces a method, Tree-Instruct, which enhances instructional complexity while simultaneously constraining thematic consistency to mitigate variations in diversity. 
Our experiments preliminarily establish a scaling law regarding complexity, show that the improvement resulting from increased complexity isn't solely due to the introduction of more training tokens and illustrate that LLMs only require complex samples for instruction tuning, rather than simple samples serving as foundational padding for curriculum learning.

\section{Tree-Instruct}
Enhancing the complexity of natural language text seems like a straightforward task for proficient LLMs. For instance, WizardLM utilizes a simple text prompt to complexify instructions as mentioned in Sec.~\ref{sec:intro}.
However, due to the extensive pre-training of LLMs on massive corpora, where models predict the next token based on the preceding context, we've noticed that LLMs can often exploit the given instruction by simply continuing the text beyond the initial prompt to artificially amplify complexity.
While adding continuation constraints can enhance the complexity of instructions, it simultaneously leads them away from the core thematic focus. 
This divergence expands the topic and domain, fostering diversity that hinders our ability to solely assess the impact of increased instruction complexity.
We leverage GPT-4 to automatically score the consistency (range in $0\sim1$) of the instructions before and after implementing in-depth deepening following WizardLM. We found that it only gets a 0.56 alignment score.
Furthermore, upon iteratively enhancing the instruction's complexity, the guidance might become ineffective, losing its original essence. For instance, it might cease to present a question, rendering it arduous for the LLM to generate a suitable response. 
This phenomenon matches with observations made by WizardLM, which prompts them to introduce the Elimination Evolving procedure.

To address this issue, we first consider \textit{what determines the complexity of natural language text}.
In linguistics and education, there is a lack of precise scientific consensus on determining the complexity of the text. No single source can precisely summarize a text's complexity.
Currently, a widely accepted perspective suggests that qualitative measures of text complexity require an informed judgment of text difficulty based on various factors. 
The standards use factors like purpose, levels of meaning, structure, language conventions, clarity, and knowledge demands to measure text difficulty~\footnote{\url{https://www.generationready.com/wp-content/uploads/2021/04/Beginners-Guide-to-Text-Complexity.pdf}}.
Among these, text structure is a more measurable indicator, as we can convert text sequences into tree structures using mature dependency or semantic tree parsers~\cite{solovyev2019computing}. Tree structures, prevalent in natural language representations, offer structural insights reflecting human text comprehension~\cite{hancke2012readability}.
Furthermore, we can gauge text complexity accurately by measuring the width and depth of trees, as a deeper and wider grammar tree signifies more intricate sentence structures~\cite{chevalier2007structural,wang2013deep}.

\begin{figure*}
      \centering
      \includegraphics[width=1\textwidth]{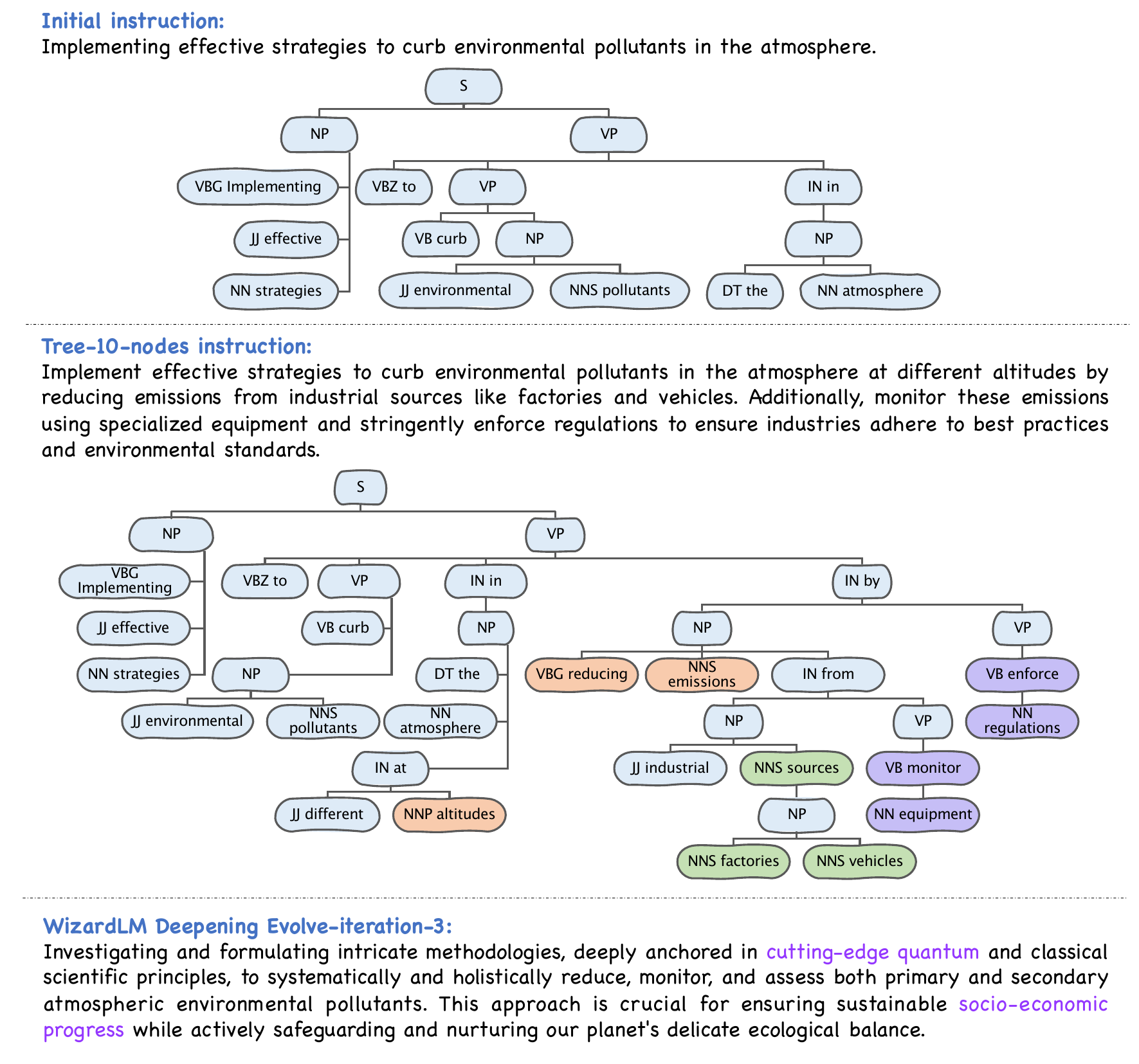}
      \caption{The instruction generated by different evolving methods: Tree-instruction after adding ten nodes and WizardLM by iteratively \textit{deepening} three times. We also demonstrate how Tree-Instruct enhances the complexity of the original instruction's semantic tree by introducing three nodes (orange), six nodes (green), and ten nodes (purple).
            \label{fig:inst}
}
  \end{figure*}

Inspired by the concept of tree complexity, we propose Tree-Instruct, wherein LLMs directly add a specific number of nodes to the semantic tree of an instruction. This increases the tree's width and depth, thereby enhancing text structure complexity.
In detail, Tree-Instruct encompasses three steps:

\textbf{Step 1: Tree Construction} involves semantic parsing, where a structured representation is created from a natural language sentence. This process yields a tree structure for an instruction. For instance, given the instruction ``\textsl{Implementing effective strategies to curb environmental pollutants in the atmosphere}'', we derive an original tree structure Tree-1 as shown in the first tree of Fig.~\ref{fig:inst}.

\textbf{Step 2: Nodes Expansion} operates on the acquired tree structure, expanding it in depth or width by adding new nodes, thus influencing the new tree's complexity. We only add meaningful nodes representing nouns or verbs, since words like adjectives or prepositions contribute little to tree complexity. The second tree in Fig.~\ref{fig:inst} illustrates Tree-2 after adding ten nodes.

\textbf{Step 3: Tree Sentenceization} aims to make LLMs revert the complex new tree structure (Tree-2) back to fluent natural language instruction by introducing connected words.
  
Additionally, we present all three steps into a single prompt, guiding LLMs to implement our requirements step by step without external semantic parsing tools (see Block~\ref{tree_prompt}, where ``\texttt{your\_added\_number}'' indicates the desired number of nodes we aim to add to the tree.) Especially we directly control the complexity by adjusting ``\texttt{your\_added\_number}''.
Visually, with more nodes added, the tree and the instruction become more complex. 
This gradual increase results in a tree with 3, 6, or 10 additional nodes, progressively increasing the complexity of instructions, as shown in Fig.~\ref{fig:inst}.
We also observe that adding nodes to the semantic tree constructs a framework for the original instruction. This approach prevents significant deviations from the main topic. 
GPT-4's automatic assessment shows that our prompt modifications maintain thematic consistency with a score 0.69.

\begin{myblock}
\label{tree_prompt}
  You are an instruction rewriter. You need to rewrite a given user instruction following Procedures step by step. You MUST ONLY return the NEW instruction you rewrite.\\

  Procedure:\\
  step-1: Parse the old ``instruction'' to a TREE-1 through Semantic Parsing in the natural language processing field.\\
  step-2: EXPAND the above NEW TREE-1 from depth or width by adding ``\textit{your\_added\_number}'' meaningful NEW Nodes as nouns or verbs to form a NEW TREE-2. The new nodes should be constructed with detailed and pertinent information.\\
  step-3: Generate a totally NEW ``instruction'' based on the expanded NEW TREE-2. \\
  
  Old instruction: ``{{\textit{your\_instruction}}}''\\
  
  New instruction: 
  \end{myblock}

\section{Experiments}

In this experiment, our primary objective is to address four key research questions:
(1) Can Tree-Instruct, compared to WizardLM's in-depth evolving, better maintain thematic consistency while augmenting complexity?
(2) Does increasing the complexity of instructions through Tree-Instruct result in a greater unleashing of LLM's latent potential, i.e., will more intricate instructions yield better outcomes?
(3) Given the same token constraints, which approach is better suited for instruction tuning: employing complex yet limited instruction data or opting for simpler but more diverse instructions?
(4) Can curriculum-based instruction tuning methods (from simpler to more complex instruction data) yield improvements similar to the substantial enhancements observed in many previous NLP tasks?

Our primary experiments are conducted on Alpaca GPT-4 dataset~\cite{peng2023instruction}, which contains a dataset of 52,000 instruction-following examples responded to by GPT-4 using prompts in Alpaca~\cite{taori2023stanford}. 
Following LIMA \citep{zhou2023lima}, we randomly select 1,000 instruction samples to form Alpaca-1K, serving as the starting point for our evolutionary process.
We query $\mathtt{gpt}$-$\mathtt{4}$~\cite{openai2023gpt} to execute Tree-Instruct process, thereby increasing the complexity of each instruction within Alpaca-1K. 
In order to analyze the scaling law, we introduce three levels of complexity by augmenting the instructions by adding 3, 6, and 10 additional nodes to the semantic tree of original instructions, respectively. 
This allows us to observe the impact of these varying complexities on the outcomes.
For the modified instructions, we employed $\mathtt{gpt}$-$\mathtt{4}$ once again to generate corresponding responses. 
To validate our findings, we replicate the results by applying the in-depth evolving with deepening prompt provided by WizardLM to the same Alpaca-1K instructions.

To demonstrate the scalability of our discoveries to larger datasets, we also conduct experiments on the extensive OpenChat dataset, which comprises 6,206 conversations between humans and GPT4, filtered from ShareGPT~\cite{openchat}. 
We employ the pre-trained LLaMA2~\cite{touvron2023llama} model as the initialization, fine-tuning it on instruction-tuning datasets generated through different methods. Each GPU processes batches of size 2 (for OpenChat evolved data, the batch size is set to 8), and the maximum sequence length was set to 4096.
For optimization, we adopt the AdamW~\cite{loshchilov2017decoupled} optimizer with a learning rate of 1e-4 and a weight decay of 0.1, following the practices established by OpenChat. 
Using Deepspeed ZeRO-2 stage and 8 A100 GPUs, we train LLaMA2 for 10 epochs on Alpaca-1K and 5 epochs on Openchat-6K data.
During inference, a temperature of 0.7 and a top-p value of 0.9 are employed to evaluate all the methods under comparison.

\paragraph{Evaluation Benchmarks}
We mainly evaluate competing methods on two benchmarks: AlpacaEval and MT-Bench \citep{zheng2023judging}. 
AlpacaEval is an LLM-based automatic evaluation, comprising 805 diverse samples, each showcasing various abilities.
In AlpacaEval, responses from different LLM methods are then compared to those from \texttt{text-davinci003} by GPT-4 auto-annotator. 
MT-Bench consists of 80 high-quality multi-turn questions to test multi-turn conversation and instruction-following ability, covering 8 common categories.
\begin{table*}
\small
    \centering
    \begin{tabular}{lccccc|l|l}
    \toprule
    Method & helpful-base & self-instruction & oasst & koala & vicuna & Alpaca-Eval & MT-Bench \\
    \midrule
Alpaca-1K & 57.36 & 38.49 & 53.72 & 37.18 & 50.00& 46.02 & 4.74 \\
Tree-3-nodes & 65.89 & 50.79 & 63.29 & 60.25 & 65.00 & 59.53 \textcolor{blue}{(+13.51)} & 6.10 \textcolor{blue}{(+1.36)}\\
Tree-6-nodes & 75.19 & 52.78 & 72.87 & 66.67 & 77.50& 66.38 \textcolor{blue}{(+20.34)} & 6.27 \textcolor{blue}{(+1.53)}\\
Tree-10-nodes & 83.72 & 58.73 & 73.43 & 75.00 & 88.75 & 72.66 \textcolor{blue}{(+26.64)}& 6.43 \textcolor{blue}{(+1.68)}\\
    \bottomrule
    \end{tabular}
    \caption{\label{tab:res_scaling}Analysis of the relationship between instruction complexity and LLM's ability. The numbers represent the win rates vs. $\mathtt{text}$-$\mathtt{davinci003}$ on various subsets of AlpacaEval and MT-Bench scores.}
\end{table*}
\subsection{Tree-Instruct is Better for Instruction Complexity Evolution}
\label{sec:consist}
\begin{figure}[h]
  \centering
  \includegraphics[width=0.48\textwidth]{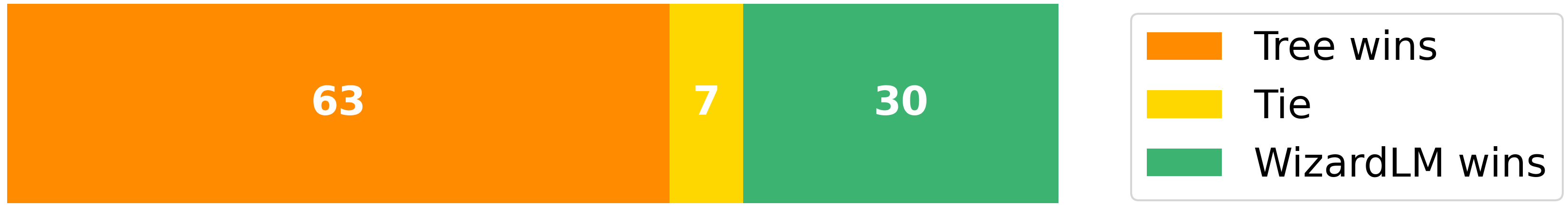}
    \caption{$\mathtt{GPT}$-$\mathtt{4}$'s comparison for the consistency preservation between Tree-6-Nodes and WizardLM's in-depth deepening with respect to the original instructions.}
    \label{fig:consistency}
\end{figure}

We start by investigating whether operating on a tree, as opposed to a sequence, better aligns with the intended objectives of the original instruction. 
Recent studies have introduced the LLMs-as-evaluator paradigm, leveraging LLMs to assess candidate samples, which closely approximates human evaluative agreement~\cite{chen2023exploring, fu2023gptscore,ji2023exploring,zhang2023wider}.
Consequently, we employ $\mathtt{gpt}$-$\mathtt{4}$ to gauge which approach exhibits greater consistency with the initial instructions. 
As depicted in Figure~\ref{fig:consistency}, the result indicates that employing Tree-Instruct, which entails adding instructions with 6 additional nodes, achieves a higher degree of alignment with the original instructions in 63\% of cases, compared to WizardLM's in-depth deepening that undergoes modifications and generates instructions with similar token quantities to Tree-6-nodes. 
This observation serves as evidence that the presence of a tree structure constraint enables LLMs to more effectively modify instructions within the framework of the original guidance, rather than diverging and incorporating unrelated content. A case study in Fig.~\ref{fig:inst} also indicates that WizardLM might produce phrases deviated from the original instruction during iterative evolution.

Furthermore, our findings demonstrate that Tree-Instruct is more effective than in-depth evolving in eliciting the capabilities of LLMs. 
We conduct evaluations on the AlpacaEval set for both methods. 
The evaluations are performed with $\mathtt{gpt}$-$\mathtt{4}$ as the evaluator, comparing the win rates of models against $\mathtt{text}$-$\mathtt{davinci003}$.
As depicted in Table~\ref{tab:result_total}, under similar total token counts, Tree-Instruct exhibits a win rate improvement of 2.2 points over WizardLM's in-depth deepening. 
We attribute this enhancement to Tree-Instruct's adeptness at closely tailoring instructions to the central topic, thereby introducing complexity without deviation. 

In contrast, in-depth evolving might deviate from the original theme and introduce irrelevant content, resulting in instructions of inadequate difficulty. 
Such instructions could potentially hinder LLMs from generating appropriate responses, rendering them less effective in the generation process.
\begin{figure}[h]
  \centering
  \includegraphics[width=0.48\textwidth]{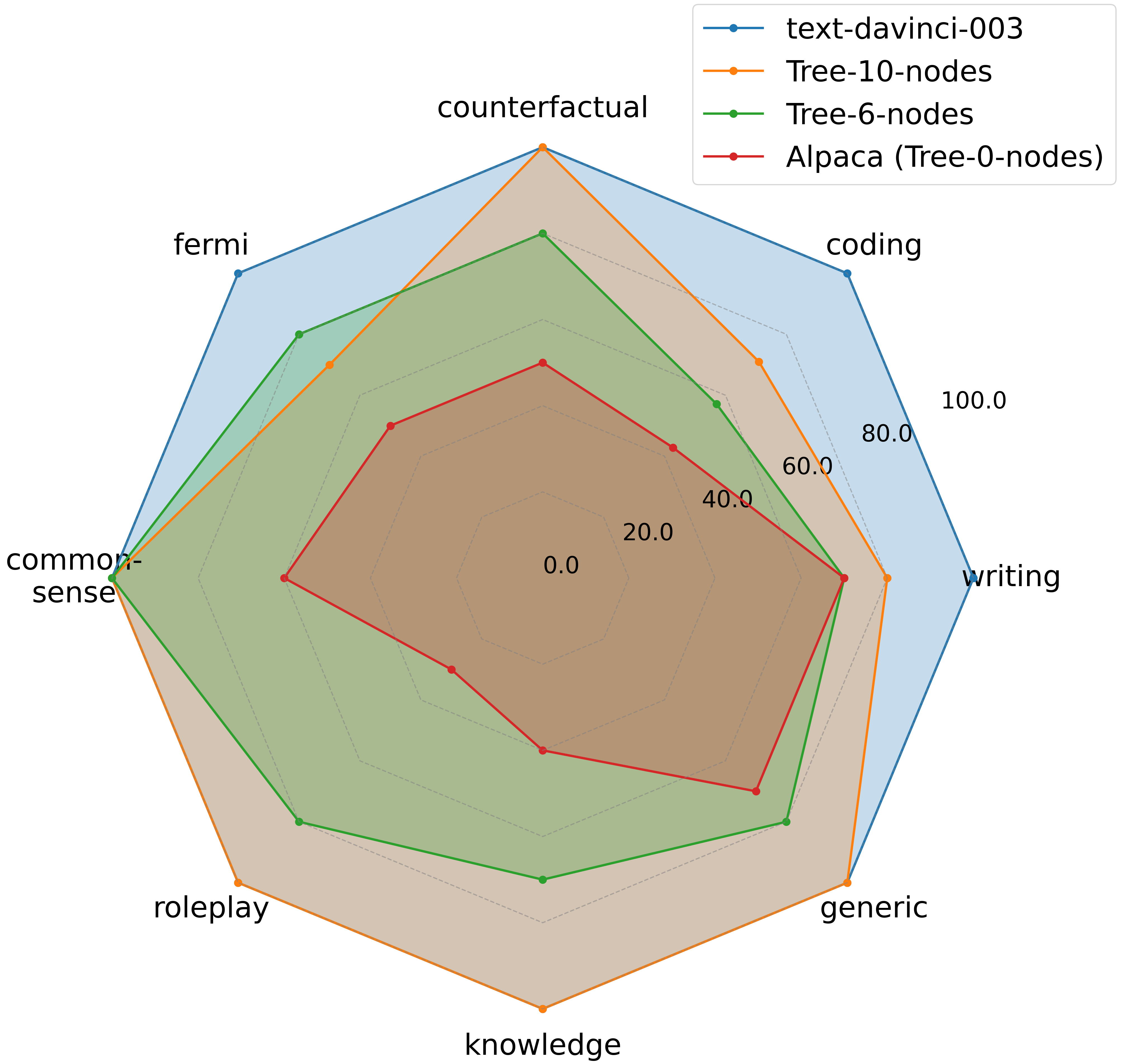}
    \caption{Evaluation of models trained on Alpaca-1K and evolved by adding various nodes vs. $\mathtt{text}$-$\mathtt{davinci003}$ on categories of the Vicuna test set.}
\end{figure}
\begin{table}
\setlength{\abovecaptionskip}{5pt}
  \centering
\small
  \begin{tabular}{lcc}
    \toprule
    \multirow{2}*{Method} & Win Rate & Token\\
    & (\%) & Length\\
    \midrule
    GPT4   & 95.28  &  1365        \\
    LLaMA2-Chat-70B & 92.66 & 1790 \\
    Claude-2 & 91.36 &  1069 \\
    OpenChat-V3.1 & 89.49 & 1484 \\
    ChatGPT & 89.37  & 827 \\
    Vicuna-33B & 88.99 & 1479 \\
    \textbf{OpenChat-LLaMA2} & 84.56  & 1730      \\
    \textbf{OpenChat-LLaMA1} & 80.87  & 1632      \\
    UltraLM-13B & 80.64 & 1087 \\
    WizardLM-13B & 75.31 & 985 \\
    \midrule
    \textbf{Tree-Instruct-LLaMA1} & 81.82 (\textcolor{blue}{+0.95})  &  1549  \\
    \textbf{Tree-Instruct-LLaMA2} & 86.19 (\textcolor{blue}{+1.63})  &  1675  \\
    \bottomrule
  \end{tabular}
  \caption{\label{tab:res_formal}Win rates of different methods vs. $\mathtt{text}$-$\mathtt{davinci003}$ on the AlpacaEval leaderboard.}
\end{table}

\begin{table}[ht]
\small
    \centering
    \begin{tabular}{lcc}
    \toprule
    Method & Win Rate (\%) & Total Token Size\\
    \midrule
    Alpaca-1K     & 46.02 & 186,464 \\
    Alpaca-4K & 55.85 & 757,042\\
    WizardLM    & 64.20 & 556,981\\ 
    \midrule
    Tree-3-Nodes & 59.53 & 385,760\\
    Tree-6-Nodes & 66.38 & 546,731\\
    Tree-10-Nodes & 72.66 & 608,556 \\
    \bottomrule
    \end{tabular}
    \caption{\label{tab:result_total}Analysis of the scaling laws for complexity and the relationship between win rate and token count with experiments based on LLaMA2.}
\end{table}

\begin{table*}[ht]
    \centering
    \begin{tabular}{lccccc|c}
    \toprule
    Method & helpful-base & self-instruction & oasst & koala & vicuna & Overall \\
    \midrule
    Mix-difficulty-training  & 73.64 & 50.79 & 68.62 & 60.26 & 70.00 & 62.59\\
    Hard-to-Easy Curriculum & 71.31 & 56.75& 66.49& 67.95&71.25 & 65.05\\
    Easy-to-Hard Curriculum & 77.52 & 57.94 & 73.93 & 74.36 & 85.00 & 70.95 \\
    \bottomrule
    \end{tabular}
    \caption{\label{tab:res_curr}Analysis of mixed difficulty training and curriculum learning. The numbers represent the win rates vs. $\mathtt{text}$-$\mathtt{davinci003}$ on various subsets of AlpacaEval.}
\end{table*}

\subsection{More Complexity, Better Capability}

After demonstrating the effectiveness of Tree-Instruct in enhancing sample complexity, we present a scaling law pertaining to complexity, as depicted in Fig.~\ref{fig:scaling_law} and Table~\ref{tab:res_scaling}.
As the number of nodes gradually increases from Tree-3-Nodes to Tree-6-Nodes and further to Tree-10-Nodes, the model's win rate on the AlpacaEval and scores on MT-Bench benchmarks exhibit a remarkable upward trend. 
This scaling law underscores the significance of complexity within instruction data.

Additionally, we carry out a meticulous evaluation for each skill/category within the Vicuna test sets. 
These sets are divided into distinct skill sets/categories, allowing for an intricate analysis of the proficiency attained through instruction tuning. 
Notably, Tree-10-Nodes outperforms Tree-6-Nodes across a majority of categories, encompassing Counterfactual, Roleplay, Knowledge, Generic, and more. 
Similar trends are evident when comparing Tree-6-Nodes with the original instructions, indicating that augmenting the complexity of Instruction data leads to a comprehensive enhancement in the capabilities of the LLM.

Finally, given that our experimentation is based on 1,000 instances, we extend our investigation to validate the effectiveness of Tree-Instruct across a larger dataset using OpenChat. OpenChat-6K is built upon 6,206 conversations between humans and GPT-4, filtered from around 90K ShareGPT conversations. It has notably achieved top rankings with much less data as an open-source LLM. 
Since OpenChat involves multi-turn conversations, we specifically complexify instructions through Tree-Instruct by adding three nodes on instructions from single-turn and several last-turn conversations. We ignore instructions that only contain generic and meaningless terms like ``stop'' or ``continue''.
The Tree-Instruct modification involves 1,147 conversations. In this process, we replace the original data with the evolved version, maintatining the same number of training samples.

As delineated in Table~\ref{tab:res_formal}, after the complexity evolution of Tree-Instruct, we enhance OpenChat's performance from 80.87\% to 81.82\% based on LLaMA1, from 84.56\% to 86.19\% on LLaMA2, underscoring the sustained efficacy of our approach across a larger volume of data.

\subsection{Less but Complex is Better Than More but Simple}
While we have demonstrated that increasing the complexity of instruction data can enhance the capabilities of LLMs, a new question arises: Is this improvement merely due to the introduction of more training tokens as complexity increases? 
Our analysis indicates that the average length of the original Alpaca data, combining both input and output, is 186 tokens. 
Upon incorporating an additional 10 nodes, this count escalates to 607 tokens – equivalent to a 3.26-fold increase in training data.
With this question in mind, we introduce a new baseline: Alpaca-4K, trained with 4,000 samples (additionally sampled 3,000 instances from the original Alpaca data). 
As shown in Table~\ref{tab:result_total}, the total token count of Alpaca-4K surpasses that of Tree-10-Nodes by 24\%. Despite this, with the same training steps, a significant 16.8\% performance gap in win rate remains. 
Compared to Alpaca-1K, there is indeed a 9.8\% improvement.
This suggests that introducing more instruction tokens does enhance model performance. Nonetheless, the effectiveness of diverse yet simple instructions still falls short compared to a smaller quantity of more complex directives.

\subsection{Curriculum Learning May Be Not Effective for Instruction Tuning}
Now, armed with three sets of data featuring increasing difficulty levels and aligned themes, we can delve into an unanswered question in instruction tuning: Is it necessary to train LLM progressively from easy to hard? 
As depicted in Table~\ref{tab:res_curr}, we embark on a trial, initially training on Tree-3-Nodes data, followed by Tree-6-Nodes, and finally Tree-10-Nodes. 
Each segment constitutes one-third of the total training steps. 
We also devise two baselines: one involving the combined training of all three difficulty levels and another wherein difficult samples are trained prior to the easy ones.

Experimental results reveal that, compared to mixed-difficulty training and training samples from hard to easy, an easy-to-hard curriculum learning approach truly enhances model performance. 
However, the performance gain from curriculum learning still slightly underperforms exclusively training on Tree-10-Nodes, the hardest dataset we construct. 
This outcome slightly contrasts with previous observations of curriculum learning. 
We attribute this variance to the fact that modern LLMs possess parameter counts several times larger than those of earlier models like BERT~\cite{bert} or T5~\cite{t5}. 
With this substantial parameter increase, LLMs are now capable of directly learning from challenging samples, diminishing the need for foundational exposure to simpler samples. The more exposure to challenging samples, the more the model's capabilities are ignited.
\section{Conclusion}
In this study, we have undertaken a preliminary exploration of the intrinsic relationship between instruction complexity and the ability of large language models to follow human instructions. We conduct extensive experiments to explore the unknown. The results reveal the following insights:
(1) As the complexity of the instruction data increases, the benefits of instruction tuning continue to amplify.
(2) The rise in complexity is partly attributed to additional tokens, yet a few intricate instructions outperform a large number of simpler instructions, all within the same token limit.
(3) A curriculum-based instruction tuning, progressing from easier to harder, might not yield the desired effectiveness; embracing increased complexity proves essential.
We anticipate that this exploration will supplement existing knowledge regarding the aspects of quality, quantity, diversity, and complexity of instruction data. This contribution aims to assist future researchers in constructing superior instruction data.

\section{Limitations}
We have conducted extensive experiments to study three unexplored aspects targeting the complexity of instruction-turning data: (1) the scaling law, (2) the impact of additional training token, and (3) curriculum learning. 
However, due to constraints in time and resources, such as computing resources or accessing GPT-4, there are still some unexplored questions: 
1. Will instruction evolution ever reach a point of convergence?
2. How can we adapt Tree-Instruct for complex tasks, such as mathematics or coding, that require intricate reasoning?
3. Are larger language models still sensitive to the complexity of instruction-tuning data?
We plan to delve into these areas in our subsequent research.

\section{Ethical Statement}
This paper studies the complexity of instruction-tuning data for large language models and proposes a novel method Tree-Instruct, to control the complexity evolution. All the datasets and models involved in this paper are publicly available. The prompts we design are also harmless and unbiased, leading to healthy and objective training data. There are no direct ethical concerns in our study.

\nocite{*}
\section{Bibliographical References}\label{sec:reference}

\bibliographystyle{lrec-coling2024-natbib}
\bibliography{lrec-coling2024-example}


\end{document}